\documentclass{article}

\usepackage{arxiv}

\usepackage[utf8]{inputenc} 
\usepackage[T1]{fontenc}    
\usepackage{hyperref}       
\usepackage{url}            
\usepackage{booktabs}       
\usepackage{amsfonts}       
\usepackage{nicefrac}       
\usepackage{microtype}      
\usepackage{lipsum}
\usepackage{graphicx}
\graphicspath{ {./Figures/} }

\usepackage{caption}
\usepackage{subcaption}
\usepackage{amsmath, amssymb}
\usepackage{braket}

\title{Neuromorphic Computing is Turing-Complete}

\author{
 Prasanna Date \\
  Computer Science \& Mathematics\\
  Oak Ridge National Laboratory\\
  Oak Ridge, TN 37932 \\
  \texttt{datepa@ornl.gov} \\
   \And
 Catherine Schuman \\
  Computer Science \& Mathematics\\
  Oak Ridge National Laboratory\\
  Oak Ridge, TN 37932 \\
  \texttt{schumancd@ornl.gov} \\
  \And
 Bill Kay \\
  Computer Science \& Mathematics\\
  Oak Ridge National Laboratory\\
  Oak Ridge, TN 37932 \\
  \texttt{kaybw@ornl.gov} \\
  \And
 Thomas Potok \\
  Computer Science \& Mathematics\\
  Oak Ridge National Laboratory\\
  Oak Ridge, TN 37932 \\
   \texttt{potokte@ornl.gov} \\
}

\begin{document}
\maketitle
\begin{abstract}
Neuromorphic computing is a non-von Neumann computing paradigm that performs computation by emulating the human brain. Neuromorphic systems are extremely energy-efficient and known to consume thousands of times less power than CPUs and GPUs. They have the potential to drive critical use cases such as autonomous vehicles, edge computing and internet of things in the future. For this reason, they are sought to be an indispensable part of the future computing landscape. Neuromorphic systems are mainly used for spike-based machine learning applications, although there are some non-machine learning applications in graph theory, differential equations, and spike-based simulations. These applications suggest that neuromorphic computing might be capable of general-purpose computing. However, general-purpose computability of neuromorphic computing has not been established yet. In this work, we prove that neuromorphic computing is Turing-complete and therefore capable of general-purpose computing. Specifically, we present a model of neuromorphic computing, with just two neuron parameters (threshold and leak), and two synaptic parameters (weight and delay). We devise neuromorphic circuits for computing all the $\mathbf{\mu}$-recursive functions (i.e., constant, successor and projection functions) and all the $\mathbf{\mu}$-recursive operators (i.e., composition, primitive recursion and minimization operators). Given that the $\mathbf{\mu}$-recursive functions and operators are precisely the ones that can be computed using a Turing machine, this work establishes the Turing-completeness of neuromorphic computing. 
\end{abstract}


\section{Introduction}
\label{sec:intro}

With the impending end of Moore's law and Dennard scaling, the computing community is looking for alternatives to conventional computing approaches such as novel materials, devices, and architectures~\cite{shalf2015computing}. Neuromorphic computing is a compelling technology in this regard \cite{mead1990neuromorphic}.
Neuromorphic computing is a non-von Neumann, brain-inspired computing paradigm that is extremely low power and can be implemented using a myriad of devices and materials, including CMOS~\cite{schuman2017survey}.
For certain applications, neuromorphic systems are faster as well as more power efficient than CPUs and GPUs~\cite{blouw2019benchmarking}. 
Over the years, a number of analog, digital and mixed analog-digital implementations of neuromorphic processors have been realized \cite{solomon2019analog,furber2014spinnaker,schemmel2010wafer,merolla2014million,davies2018loihi,bauer2019real}.
Because of their low power nature, neuromorphic computers are expected to shine in critical use cases such as autonomous vehicles, robotics, edge computing, internet of things, and wearable technologies~\cite{mitchell2017neon,milde2017obstacle,park2020selecting,ham2020one,li2020flexible}. 



Neuromorphic computers were originally developed to perform  spike-based neural network-style computation, and primarily used for machine learning and computational neuroscience applications \cite{schuman2020evolutionary,date2018efficient}.
However, they have many characteristics that make them attractive to other application areas.  
For instance, neuromorphic computers are inherently massively parallel, scalable, have co-located processing and memory, and can perform event-driven, asynchronous computation.  
In the recent years, neuromorphic approaches have been used to address problems in graph theory, partial differential equations, constraint satisfaction optimization, and spike-based simulations~\cite{aimone2020provable, schuman2019shortest,kay2020neuromorphic,hamilton2020spike,smith2020solving,fonseca2017using, yakopcic2020solving,aimone2018non,hamilton2020modeling}.  

The demonstration of broader applicability of neuromorphic computing begs the question:
\emph{Can neuromorphic computing be used for general-purpose computing, i.e., is it Turing-complete?} 
The idea that neuromorphic computing is Turing-complete is loosely held within the community.  
However, its Turing-completeness has never been proven.
To this extent, our main contributions in this paper are:
\begin{enumerate}
    \item We propose a simple model of neuromorphic computing, where neurons and synapses have two parameters each. 
    \item We devise neuromorphic circuits to compute all the $\mu$-recursive functions and operators. In doing so, we show our model is Turing-complete.
\end{enumerate}

\section{Related Work}
\label{sec:related}




The Turing machine, proposed by Alan Turing in 1937, is a model of computation that can express any arbitrary computation \cite{turing1937computability}. 
The von Neumann architecture used in today's computers is a physical manifestation of the Turing machine~\cite{von1993first,turing2009computing}.
Contemporary to the Turing machine, G\"{o}del and Herbrand introduced another model of computation, called the $\mu$-recursive functions \cite{godel1934undecidable,smith2013introduction}.
It was subsequently shown that these two models were equivalent \cite{turing1937computability}. 
A computation model is said to be Turing-complete if it can compute the same set of functions as the Turing machine \cite{church1936unsolvable,church1936note,church1940formulation}.
Since the set of $\mu$-recursive functions is precisely that set, it suffices to show that a computation model can compute all the $\mu$-recursive functions in order to show that it is Turing-complete.


There have been some efforts to show certain models of of spiking and recurrent neural networks are Turing-complete. 
Maass shows that spiking neural networks with synaptic plasticity are Turing complete \cite{maass1996lower}.
Siegelmann shows that analog neural networks---which are equivalent to Maass' spiking neural networks with synaptic plasticity---allow for super-Turing compute power~\cite{siegelmann2003neural}.
Cabessa and Siegelmann show that recurrent neural networks have super-Turing capabilities when synaptic plasticity is assumed~\cite{cabessa2014super}. 
Cabessa shows that a rational-weighted recurrent neural network employing spike-timing-dependent-plasticity (STDP) is Turing-complete \cite{cabessa2019turing}.

All of the above models focus on either spiking or recurrent neural networks.
Some assume more complex neuron and synapse models.
Most importantly, synaptic plasticity is critical to all of their arguments. 
Their results are restricted to neuromorphic computing models with synaptic plasticity and are not generalizable. 
There are Turing-completeness claims for general neuromorphic systems, as well as a notion of `neuromorphic completeness' \cite{kazempour2015seminar,zhang2020system,rhodes2020brain}.
However, they are not supported by formal proofs and do not establish the Turing-completeness of neuromorphic computing. 

The need for a computability and complexity theory for neuromorphic computing is pressing, especially in the context of post-Moore computing. 
Neuromorphic computers, due to their low energy consumption, are prime candidates for co-processors and accelerators in extremely heterogeneous computing systems~\cite{liu2014heterogeneous,aimone2018non}. 
However, because of the absence of a sound computability and complexity theory for neuromorphic computing~\cite{kwisthout2020computational}, we are faced with a murky understanding of the \emph{big}-$\mathcal{O}$ runtimes~\cite{knuth1976big} of neuromorphic algorithms and unable to compare them to their conventional counterparts.
This is not the case for other post-Moore computing paradigms such as quantum computing \cite{vetter2017architectures,nielsen2002quantum,date2019efficiently}, which has a well-founded literature on the theory of quantum computation ~\cite{steane1998quantum,bernstein1997quantum}.
As a result, it is possible to objectively compare quantum algorithms to their classical counterparts \cite{arute2019quantum}.
In this paper, we take the first steps to establish the computability and complexity of neuromorphic computing.

\section{Neuromorphic Computing Model}
\label{sec:model}

\begin{figure}[t!]
    \centering
    \begin{subfigure}[b]{0.5\textwidth}
        \centering
        \includegraphics[scale=0.4,trim=200px 210px 270px 230px, clip]{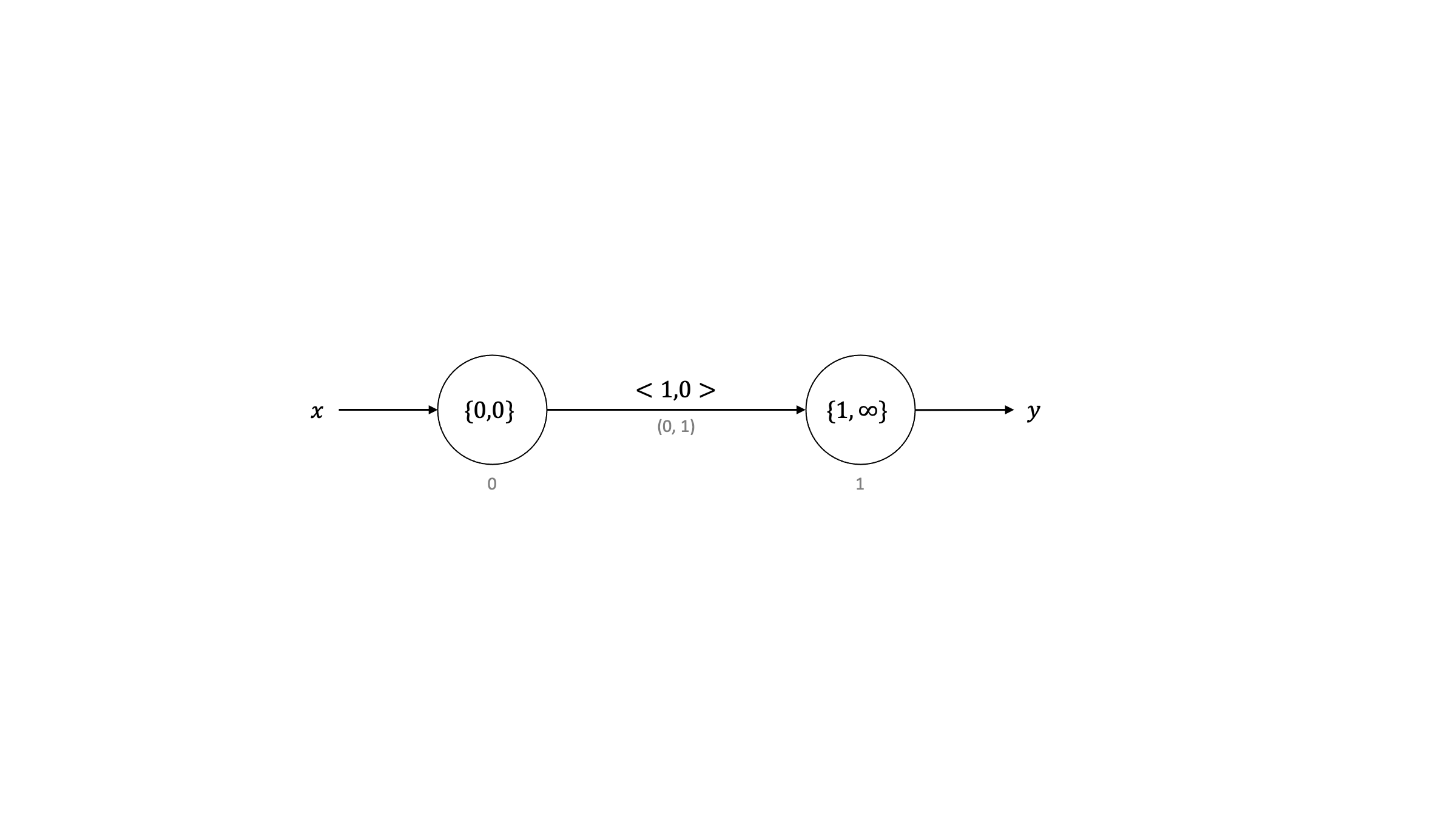}
        \caption{Symbolic notation describing a neuromorphic circuit.}
        \label{fig:notation}
    \end{subfigure}\\
    \begin{subfigure}[b]{0.5\textwidth}
        \centering
        \includegraphics[scale=0.4,trim=400px 210px 400px 120px, clip]{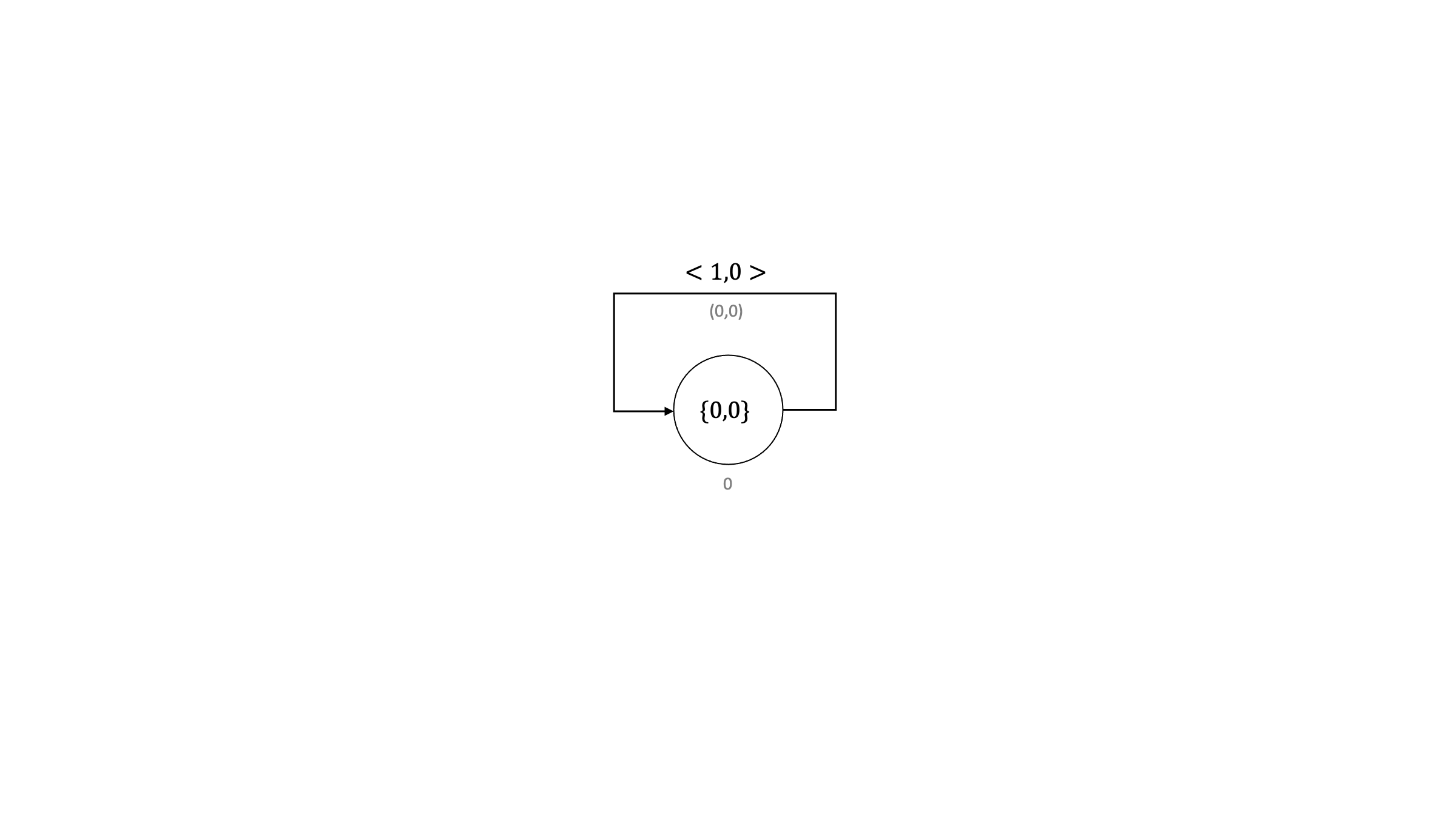}
        \caption{Self synapse \texttt{(0, 0)} going from neuron \texttt{0} to itself.}
        \label{fig:self-synapse}
    \end{subfigure}\\
    \begin{subfigure}[b]{0.5\textwidth}
        \centering
        \includegraphics[scale=0.4,trim=200px 110px 270px 50px, clip]{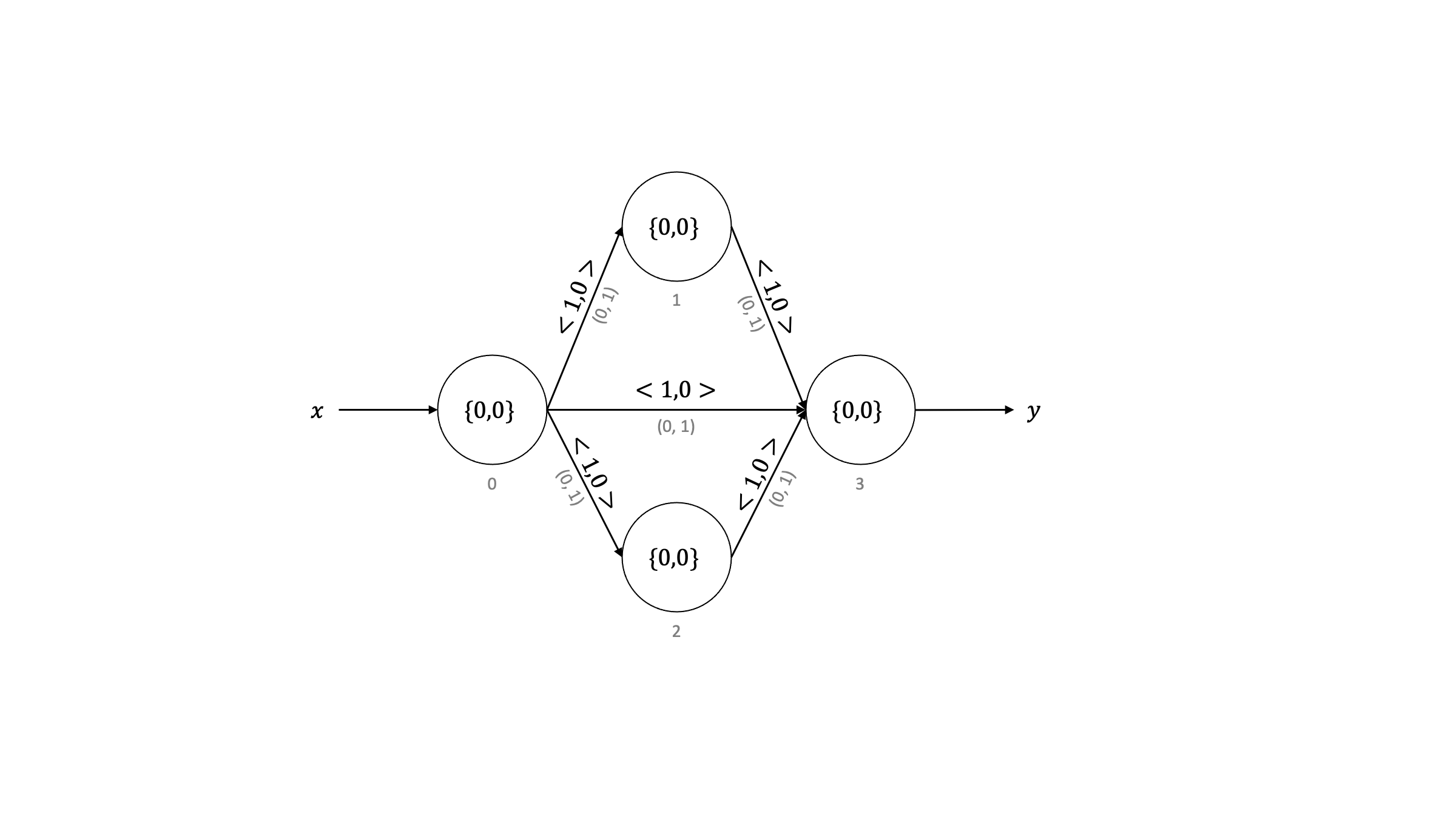}
        \caption{Multiple synapses from neuron \texttt{0} to neuron \texttt{3} via proxy neurons.}
        \label{fig:multiple-synapses}
    \end{subfigure}
    \caption{Illustration of the symbolic notation, self-synapse and multiple synapses.}
    \label{fig:notation-self-synapse-and-multiple-synapses}
\end{figure}





\subsection{Structural Components}
\label{sub:structural-components}
Our model of neuromorphic computing is comprised of two structural components: neurons and synapses.
A neuron is the fundamental unit from which a neuromorphic circuit is built.
A synapse constitutes a connection from a pre-synaptic (source) neuron to a post-synaptic (target) neuron.
We assume the existence of infinitely many neurons and synapses.
This is essential to model the inherent parallelism of neuromorphic systems.
We further assume that all-to-all connectivity between neurons is possible. 
Neurons could also receive signals from, and send signals to external synapses, which are used for input/output (I/O) of data.
Such neurons are called I/O neurons.

Figure \ref{fig:notation} shows the symbolic notation for neurons, synapses and neuromorphic circuits throughout this paper.
The circles represent neurons and arrows represent synapses.
Neurons are referenced using whole numbers shown below them.
Synapses are referenced using a tuple containing the references of the pre-synaptic neuron and the post-synaptic neuron.
External synapses do not have explicit references.
Figure \ref{fig:notation} shows an external synapse on the left feeding an input $x$ to the input neuron \texttt{0}.
Neuron \texttt{0} is connected to neuron \texttt{1} via the synapse \texttt{(0, 1)}.
Neuron \texttt{1} returns the output $y$ through the external synapse on the right.
We allow a neuron to be connected to itself via a self-synapse as shown in Figure \ref{fig:self-synapse}. 
We do not allow a pre-synaptic and a post-synaptic neuron to be directly connected via multiple synapses in order to be consistent with our referencing scheme.
Such a functionality could be achieved by routing a signal through multiple proxy neurons as shown in Figure \ref{fig:multiple-synapses}.
Throughout the paper, we will represent smaller, simpler circuits using boxes, and use them in larger, more complex circuits.
These boxed circuits will be referenced similar to neurons (see Figure \ref{fig:projection-function-circuit} for example).


\subsection{Functional Aspects and Parameters}
\label{sub:functional-aspects-and-parameters}
The neurons in our model are based on the leaky integrate-and-fire neurons \cite{lapique1907recherches,abbott1999lapicque}.
Our neurons accumulate signals from incoming synapses in their internal state until a threshold is reached. 
After reaching the threshold, they spike, and send their signal via outgoing synapses.
The neurons reset to an internal state of zero after spiking.
Our neurons have a leak, which specifies the time it takes to push their internal state back to zero in case they do not spike.
For instance, neurons with a leak of $0$ (instantaneous leak) would have no recollection of their internal state. 
On the other hand, neurons with infinite leak would remember their internal state exactly until they spike and reset.
We denote the internal state of the $i^{\text{th}}$ neuron using $v_i$, which is integer-valued. 
The two neuron parameters---threshold and leak---are whole numbers denoted by $\nu_i$ and $\lambda_i$.
We use curly parentheses to specify neuron parameters $\{\nu_i, \lambda_i\}$ in our circuit diagrams.

Synapses in our model receive signals from their pre-synaptic neuron, multiply the signal by their weight, stall for a time indicated by their delay, and deposit the signal in their post-synaptic neuron.
For a synapse connecting neuron \texttt{i} to neuron \texttt{j}, its weight and delay are denoted by $\omega_{i,j}$ and $\delta_{i,j}$.
The weights are integer-valued and the delays are whole numbers.
In our circuit diagrams, we indicate the synaptic parameters using arrow parentheses $\braket{\omega_{i,j}, \delta_{i,j}}$ on top of the synapses.
External synapses do not have weights or delays.

We assume it takes one unit of time for a spike to travel across any synapse in the absence of delay.
This assumption is fundamental to determining the computational complexity of neuromorphic algorithms.
Under this assumption, the total time taken for a spike to travel across a synapse having $\delta$ delay would be $\delta+1$.
Usually, spikes in neuromorphic systems do not have an associated value.
However, it is possible to encode integers, rational numbers and real numbers as spikes using spatio-temporal encoding schemes \cite{schuman2019non, pan2019neural, kiselev2016rate}.
For the sake of clarity, and for the ease of explaining neuromorphic circuits in Section \ref{sec:proof}, we will operate at a higher level of abstraction and let spikes have an associated integer value.

A neuromorphic algorithm is defined by the configuration of a neuromorphic circuit. 
We assume that a neuromorphic circuit interfaces with data that is also neuromorphic, i.e. encoded as spikes.
The outputs of a neuromorphic circuit are represented by the spiking behavior of the neurons. 
The spikes have an associated time at which they occur and a neuron on which they occur.  
The output of the circuit is called the spike raster, which enumerates the time and neuron for each spike that occurs over the course of the circuit's operation.  







\section{Proof of Turing Completeness}
\label{sec:proof}

It is possible to simulate neuromorphic computations exactly on von Neumann machines using simulators such as the NEST neural simulator~\cite{gewaltig2007nest}.
Since von Neumann machines are physical manifestations of the Turing machine, we know that neuromorphic computing can at most be Turing-complete.
In this section, we present neuromorphic circuits for each of the three $\mu$-recursive functions, i.e., constant function, successor function, and projection function; and each of the three $\mu$-recursive operators, i.e., composition operator, primitive recursion operator and minimization operator.
The set of $\mu$-recursive functions and operators is known to be Turing-complete \cite{turing1937computability}.
By showing that our model can compute all of the $\mu$-recursive functions and operators, we would prove that it is Turing-complete.
We let $\mathbb{N}$ be the set of natural numbers.

\subsection{Constant Function ($C_k$)}
\label{sub:constant-function}
\begin{figure*}[t!]
    \centering
    \begin{subfigure}[b]{0.4\textwidth}
        \centering
        \includegraphics[scale=0.4,trim=220px 160px 220px 160px,clip]{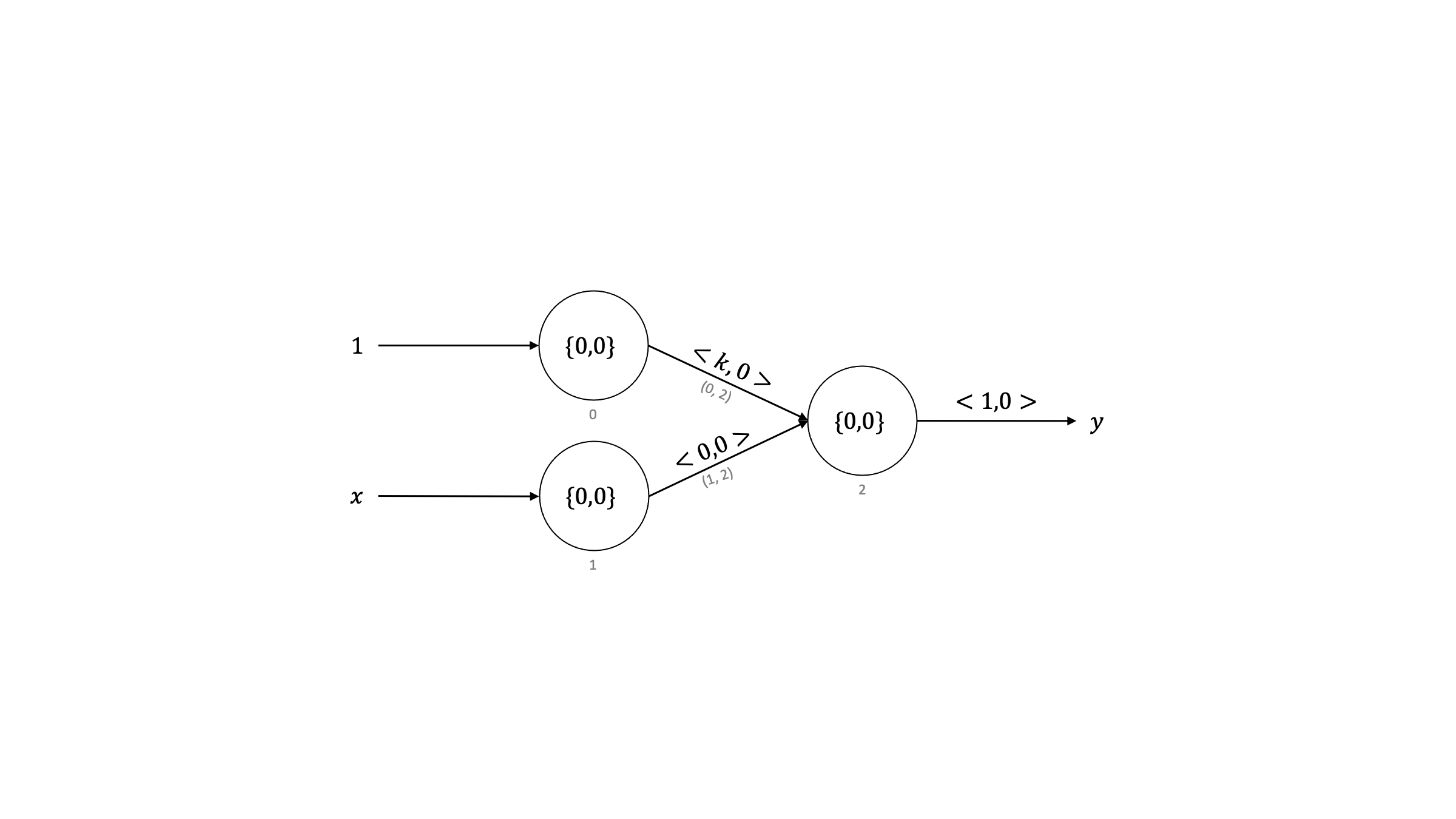}
        \caption{Constant function circuit.}
        \label{fig:constant-function-circuit}
    \end{subfigure}
    \begin{subfigure}[b]{0.4\textwidth}
        \centering
        \includegraphics[scale=0.4,trim=220px 160px 220px 160px,clip]{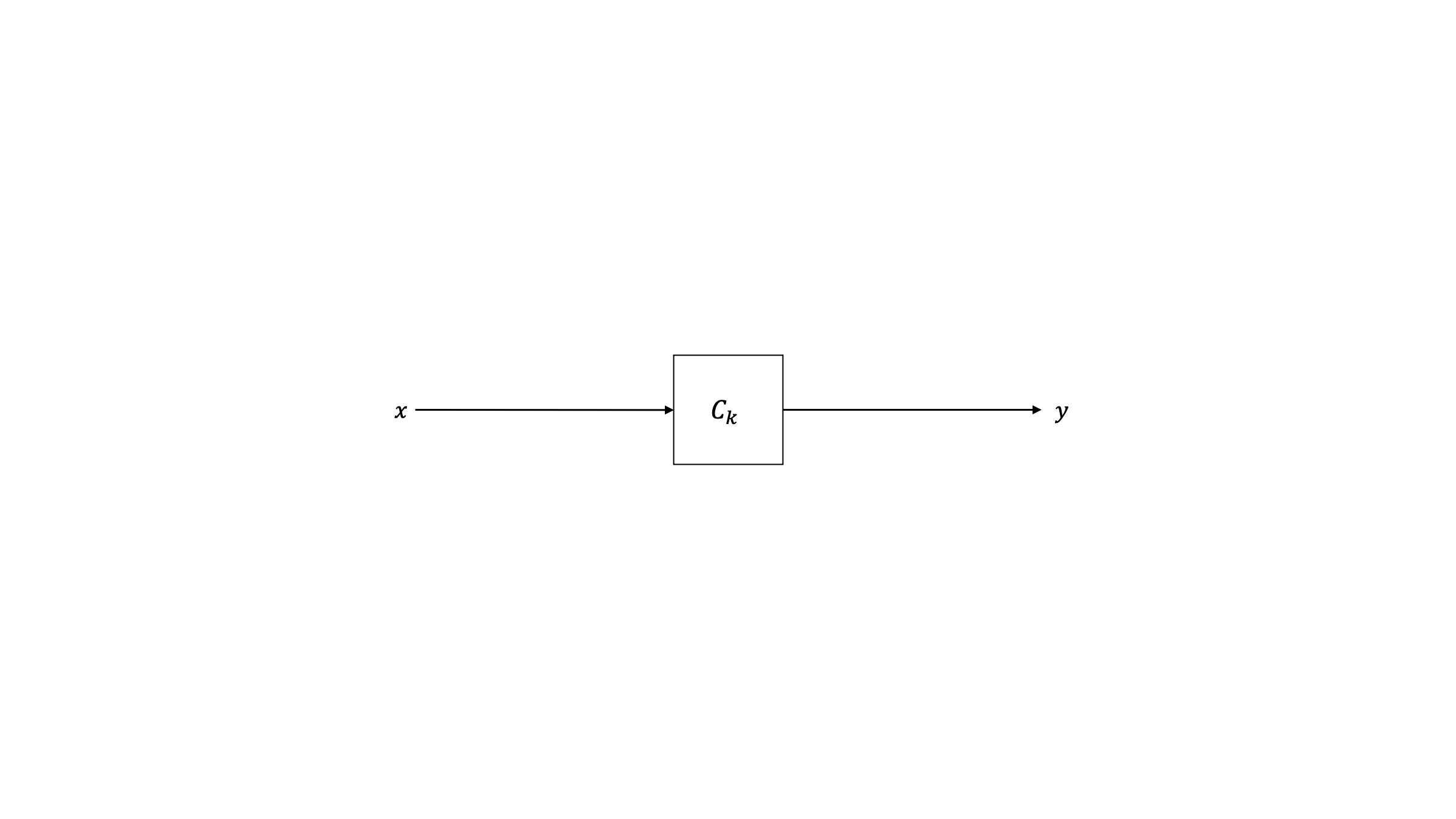}
        \caption{Constant function abstraction.}
        \label{fig:constant-function-abstraction}
    \end{subfigure} \\
    \begin{subfigure}[b]{0.4\textwidth}
        \centering
        \includegraphics[scale=0.4,trim=200px 160px 220px 120px,clip]{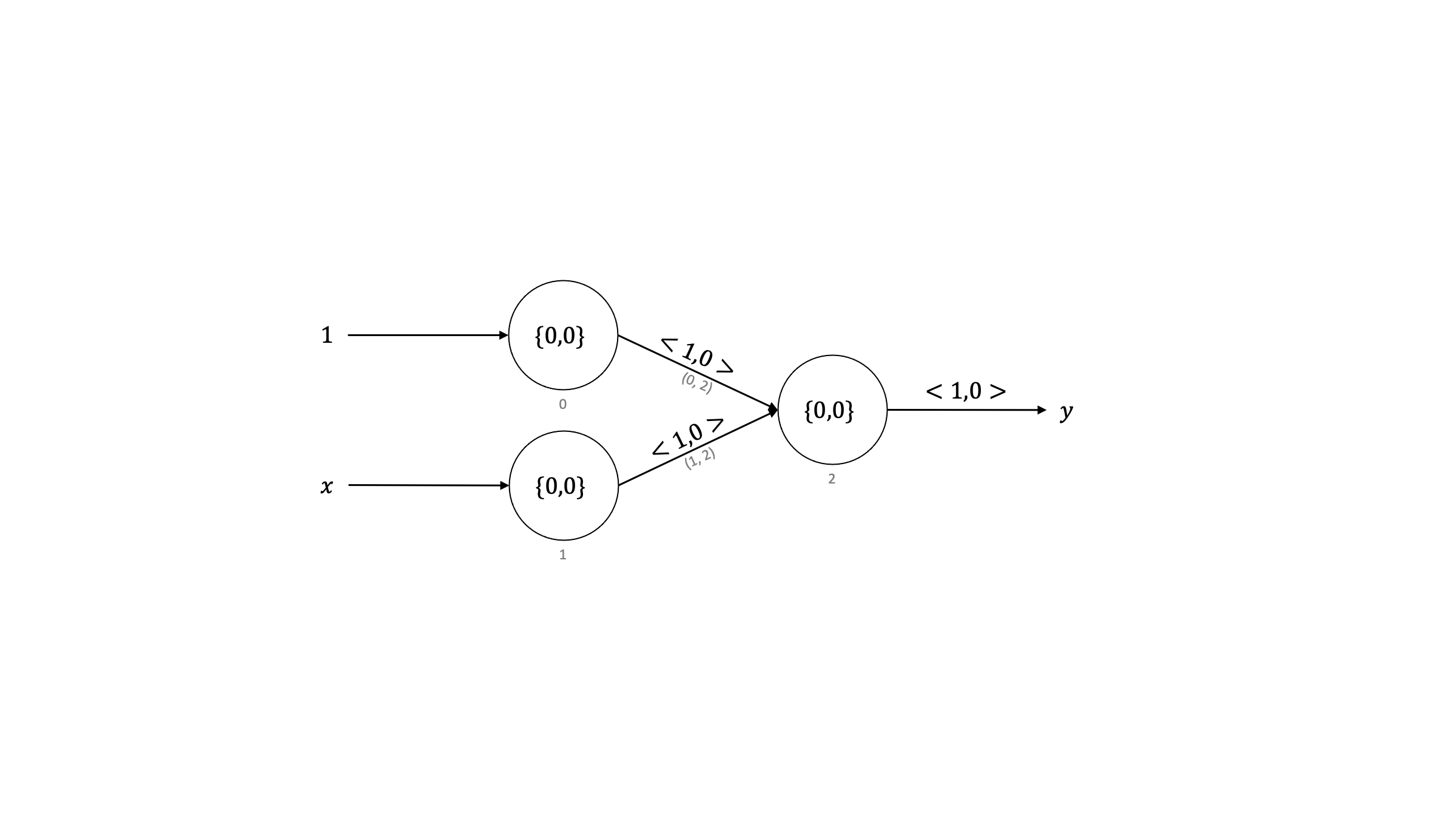}
        \caption{Successor function circuit.}
        \label{fig:successor-function-circuit}
    \end{subfigure}
    \begin{subfigure}[b]{0.4\textwidth}
        \centering
        \includegraphics[scale=0.4,trim=220px 160px 220px 120px,clip]{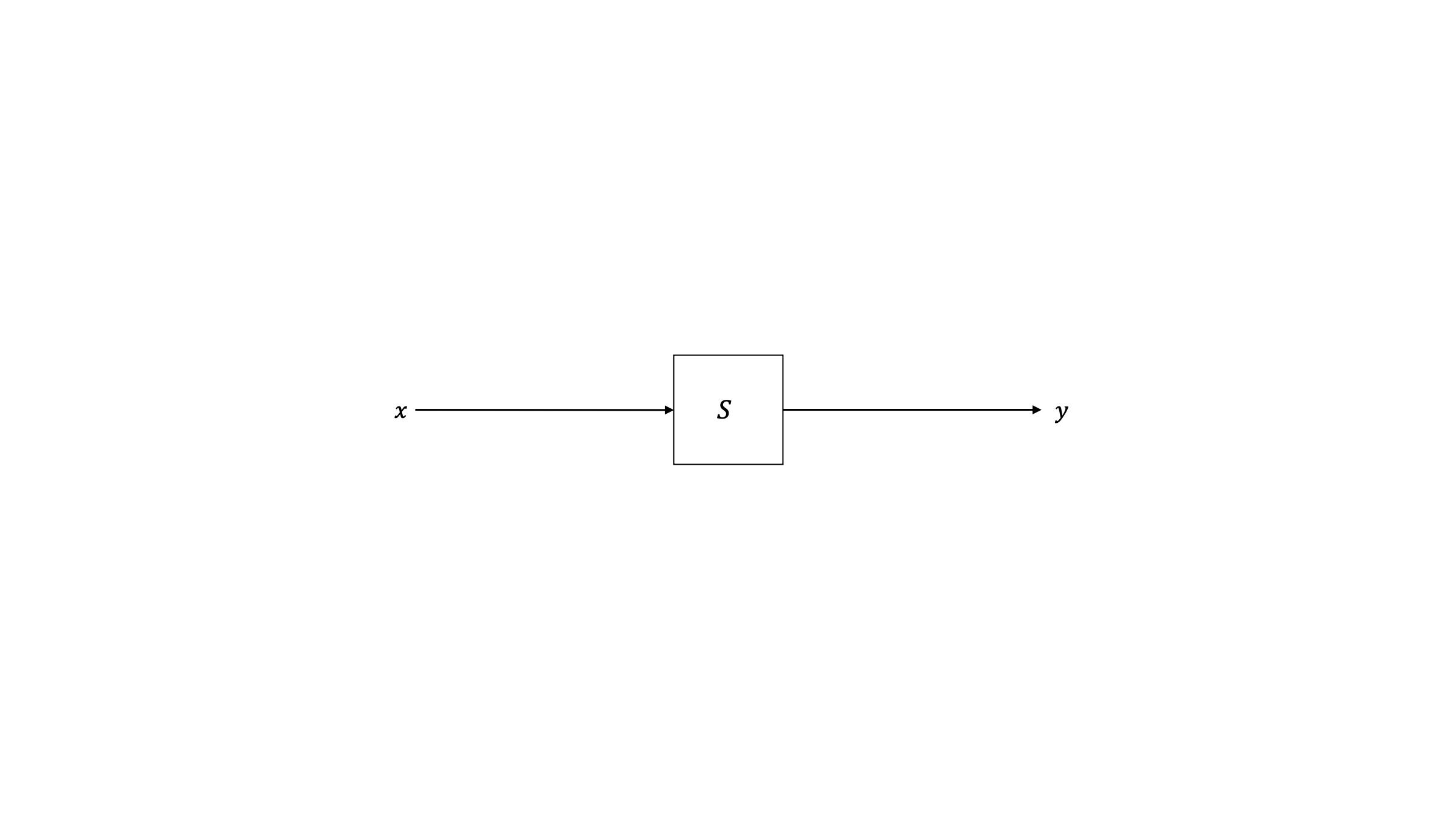}
        \caption{Successor function abstraction.}
        \label{fig:successor-function-abstraction}
    \end{subfigure}
    \label{fig:constant-successor-and-predecessor-functions}
    \caption{Neuromorphic circuits and abstractions for the constant and successor functions.}
\end{figure*}

The constant function is the first $\mu$-recursive function.
For a natural number $x$, it returns a constant natural number $k$. It is defined as:
\begin{align}
    C_k(x) := k \label{eq:constant-function-definition}
\end{align}

Figure \ref{fig:constant-function-circuit} shows the neuromorphic circuit that computes the constant function.
Neuron \texttt{1} receives the input $x$.
Parallely, neuron \texttt{0} receives the input $1$.
Both neurons \texttt{1} and \texttt{0} spike as their internal states exceed their thresholds of $0$.
The synapse \texttt{(1, 2)} multiplies the signal $x$ by its synaptic weight, which is $0$.
Parallely, synapse \texttt{(0, 2)} multiplies the signal $1$ by its synaptic weight, which is $k$.
As a result, neuron \texttt{2} receives a signal of $0$ from synapse \texttt{(1, 2)} and a signal of $k$ from synapse \texttt{(0, 2)}.
It integrates the signals and spikes as its internal state $k$ exceeds its threshold of $0$.
The output $y$ received from the circuit is $k$ as desired.
The abstraction of the constant circuit is shown in Figure \ref{fig:constant-function-abstraction}.

\subsection{Successor Function ($S$)}
\label{sub:successor-function}
For a natural number $x$, the successor function returns $x+1$.
The successor of $0$ is defined as $1$.
The successor function is defined as:
\begin{align}
    S(x) := x + 1 \label{eq:successor-function-definition}
\end{align}

Figure \ref{fig:successor-function-circuit} shows the successor function. 
It is similar to the constant function, with two differences.
The weights of both synapses \texttt{(0, 2)} and \texttt{(1, 2)} are $1$.
Neuron \texttt{2} receives $1$ from \texttt{(0, 2)} and $x$ from \texttt{(1, 2)}.
Its internal state becomes $x + 1$, and it spikes.
The output $y$ is $x + 1$ as desired.
Figure \ref{fig:successor-function-abstraction} shows the abstraction of the successor function.

\subsection{Projection Function ($P$)}
\label{sub:projection-function}
\begin{figure*}[t!]
    \centering
    \includegraphics[scale=0.6,trim=40px 25px 50px 10px,clip]{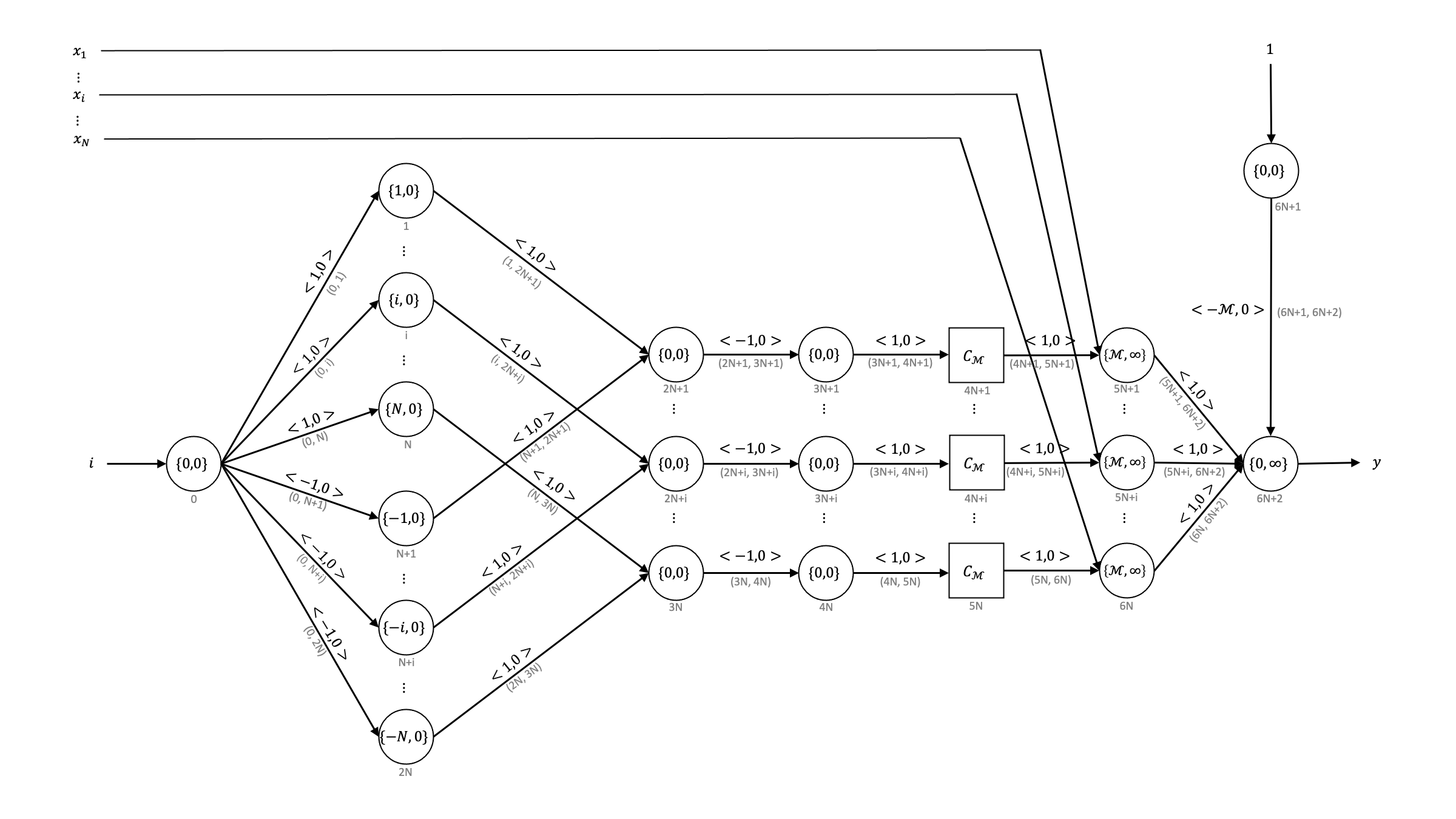}
    \caption{Neuromorphic circuit for the projection function.}
    \label{fig:projection-function-circuit}
\end{figure*}

\begin{figure}[t!]
    \centering
    \includegraphics[scale=0.5,trim=240px 190px 250px 160px,clip]{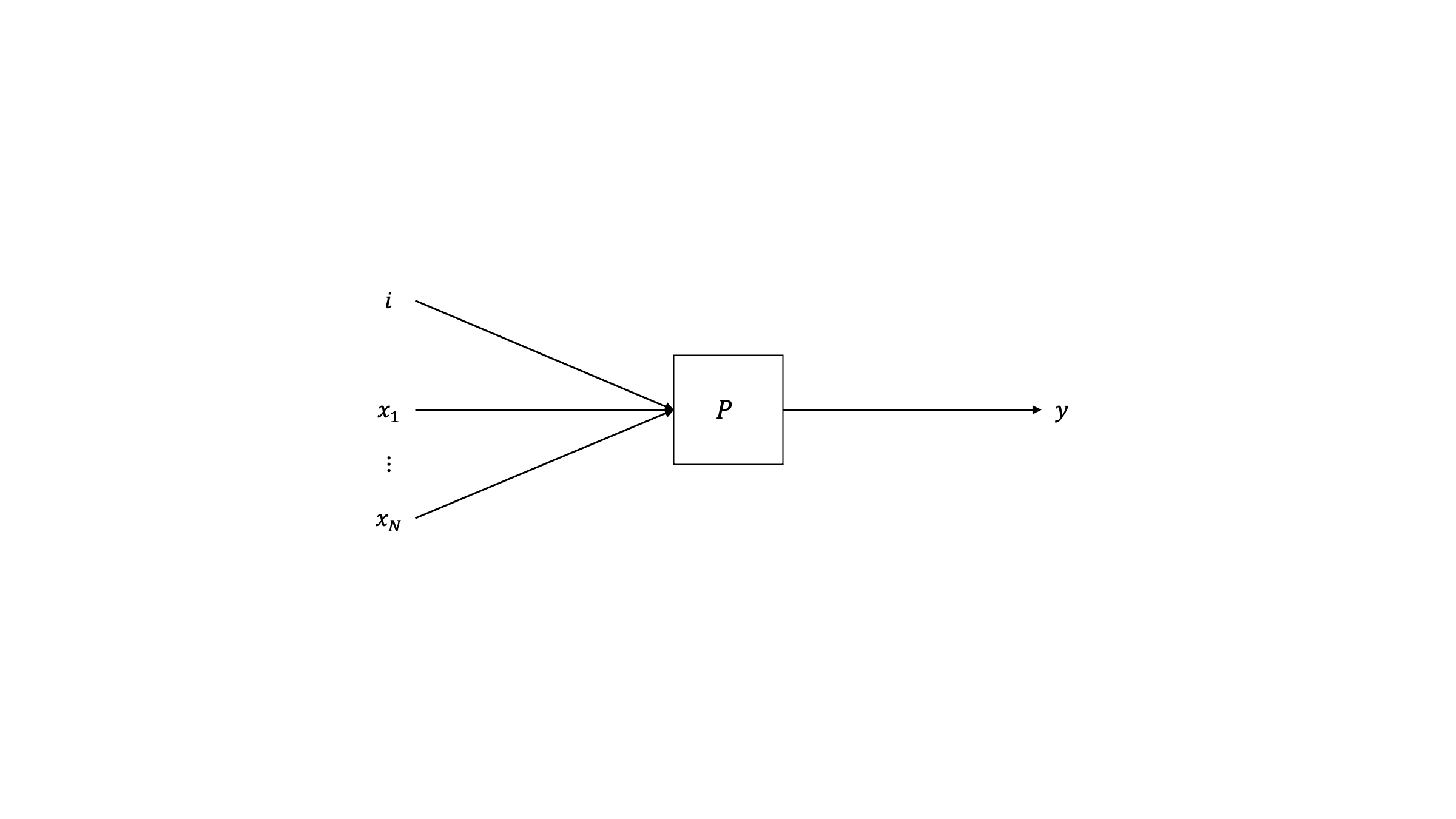}
    \caption{Abstraction for the projection function.}
    \label{fig:projection-function-abstraction}
\end{figure}

The projection function resembles getting the $i^{\text{th}}$ element of an array.
Formally, given natural numbers $i$ and $N$ such that $1 \le i \le N$, and natural numbers $x_1, \ldots, x_N$, the projection function returns $x_i$.
It is defined as:
\begin{align}
    P(i, x_1, \ldots, x_N) := x_i \label{eq:projection-function-definition}
\end{align}

The projection function and its abstraction are shown in Figures \ref{fig:projection-function-circuit} and \ref{fig:projection-function-abstraction}.
Inputs $x_1, \ldots, x_N$ are received in neurons \texttt{5N + 1} through \texttt{6N}.
These neurons have a threshold of $\mathcal{M}$, which is a large positive number, and an infinite leak.
After receiving the inputs, their internal states become $x_1, \ldots, x_N$ respectively, but they do not spike.
Parallely, neuron \texttt{6N + 1} receives an input of $1$ and spikes.
This signal is multiplied by $-\mathcal{M}$ in the synapse \texttt{(6N + 1, 6N + 2)}.
The neuron \texttt{6N + 2} has an infinite leak and stores the value of $-\mathcal{M}$ in its internal state.

Simultaneously, neuron \texttt{0} receives $i$. 
It spikes and sends $i$ to its outgoing synapses. 
Synapses \texttt{(0, 1)} through \texttt{(0, N)} multiply their signals by $1$, and send them to neurons \texttt{1} through \texttt{N}.
Synapses \texttt{(0, N + 1)} through \texttt{(0, 2N)} multiply their signals by $-1$ and send them to neurons \texttt{N + 1} through \texttt{2N}.
While the neurons \texttt{1} through \texttt{N} receive a signal of $i$, the neurons \texttt{N + 1} through \texttt{2N} receive a signal of $-i$.
The former neurons have thresholds of $1$ through $N$, and the latter neurons have thresholds of $-1$ through $-N$.
Thus, neuron \texttt{1} through \texttt{i} spike with a value of $i$ and neurons \texttt{N + i} through \texttt{2N} spike with a value of $-i$.

We now look at the neurons \texttt{2N + 1} through \texttt{3N}, which have two incoming synapses each.
Neuron \texttt{2N + 1} receives $i$ from \texttt{(1, 2N + 1)}, but does not receive anything from \texttt{(N + 1, 2N + 1)}.
Same is true for neurons \texttt{2N + 1} through \texttt{2N + i - 1}. 
All of their internal states become $i$.
The neuron \texttt{2N + i} receives $i$ and $-i$ from \texttt{(i, 2N + i)} and \texttt{(N + i, 2N + i)}, so that its internal state is zero.
The neuron \texttt{3N} receives $-i$ from \texttt{(2N, 3N)}, but does not receive anything from \texttt{(N, 3N)}.
Same is true for neurons \texttt{2N + i + 1} through \texttt{3N}.
Their internal states become $-i$.
Thus, neurons \texttt{2N + 1} through \texttt{2N + i - 1} have internal states of $i$ and spike; 
neuron \texttt{2N + i} has an internal state of $0$ and spikes; 
and, neurons \texttt{2N + i + 1} through \texttt{3N} have internal states of $-i$ and do not spike.

The signals from neurons \texttt{2N + 1} through \texttt{2N + i} are multiplied by $-1$.
Neurons \texttt{3N + 1} through \texttt{3N + i - 1} receive $-i$ and do not spike.
Neuron \texttt{3N + i} receives $0$ and spikes.
So, we have successfully isolated the signal corresponding to $x_i$.
It is passed through the constant function, which returns $\mathcal{M}$.
This value reaches the neuron \texttt{5N + i}, which has an internal state of $x_i$.
After receiving $\mathcal{M}$ from \texttt{(4N + i, 5N + i)}, its internal state becomes $\mathcal{M} + x_i$ and it spikes.
Thus, $\mathcal{M} + x_i$ reaches the neuron \texttt{6N + 2}, which has an internal state of $-\mathcal{M}$.
Upon receiving the signal from \texttt{(5N + i, 6N + 2)}, its internal state becomes $x_i$ and it spikes.
The output $y$ thus equals $x_i$ as desired.

\subsection{Composition Operator ($\circ$)}
\label{sub:composition-operator}
\begin{figure*}[t!]
    \centering
    \begin{subfigure}[b]{0.6\textwidth}
        \centering
        \includegraphics[scale=0.35,trim=105px 65px 100px 90px,clip]{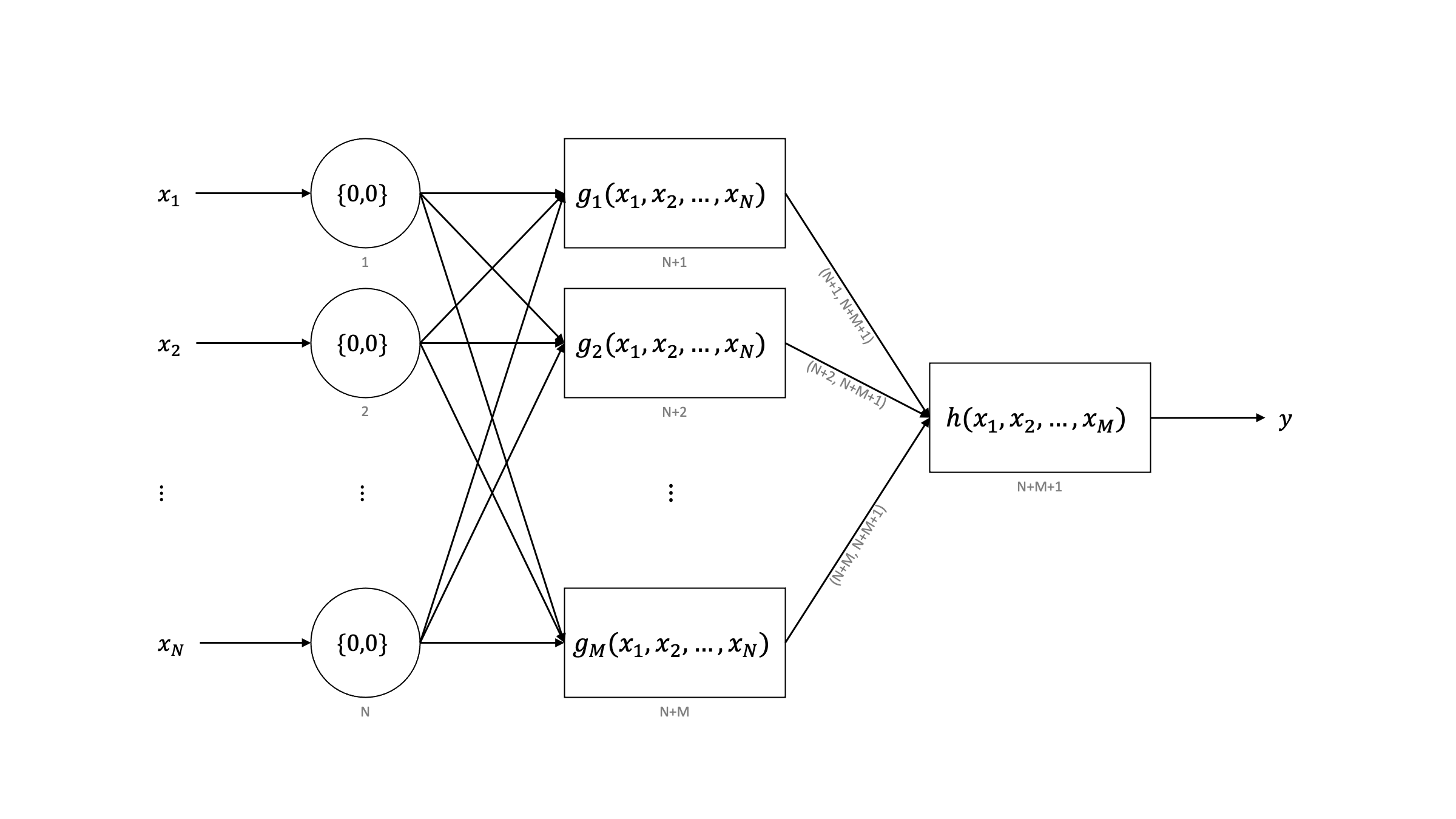}
        \caption{Composition operator circuit.}
        \label{fig:composition-operator-circuit}
    \end{subfigure}
    \begin{subfigure}[b]{0.35\textwidth}
        \centering
        \includegraphics[scale=0.35,trim=200px 150px 200px 150px,clip]{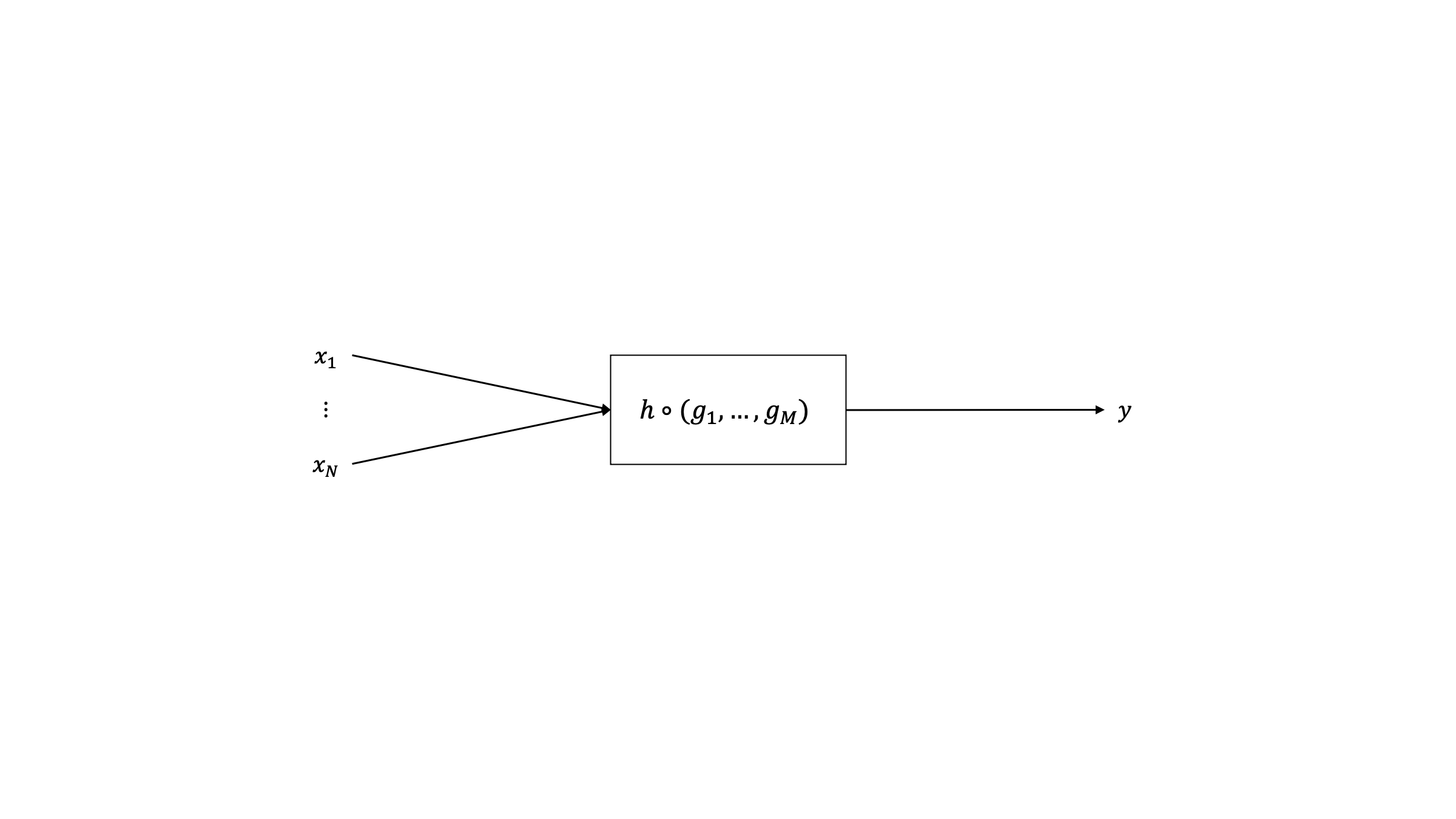}
        \caption{Composition operator abstraction.}
        \label{fig:composition-operator-abstraction}
    \end{subfigure}
    \label{fig:composition-operator}
    \caption{Neuromorphic circuit and abstraction for the composition operator. All synapses are $\braket{1, 0}$ unless explicitly stated.}
\end{figure*}

The composition operator is used for function composition in any computation model.
Let $M$ and $N$ be natural numbers.
Given the function, $h: \mathbb{N}^M \rightarrow \mathbb{N}$, and the functions, $g_1, \ldots, g_M: \mathbb{N}^N \rightarrow \mathbb{N}$, the composition operator $h \circ (g_1, \ldots, g_M)$ is defined as:
\begin{align}
    h \circ (g_1, \ldots, & g_M)(x_1, \ldots, x_N) := h(g_1(x_1, \ldots, x_N), \ldots, g_M(x_1, \ldots, x_N)) \label{eq:composition-operator-definition}
\end{align}

The composition operator and its abstraction are shown in Figures \ref{fig:composition-operator-circuit} and \ref{fig:composition-operator-abstraction}.
Neurons \texttt{1} through \texttt{N} receive $x_1, \ldots, x_N$.
They spike and distribute the inputs to functions $g_1$ through $g_M$.
Outputs from these functions are sent to the function $h$.
Output of the function $h$ is returned as the output of the circuit $y$, which equals $h \circ (g_1, \ldots, g_M)$, as desired.



\subsection{Primitive Recursion Operator ($\rho$)}
\label{sub:primitive-recursion-operator}

\begin{figure*}[t!]
    \centering
    \begin{subfigure}[b]{0.4\textwidth}
        \centering
        \includegraphics[scale=0.4,trim=200px 200px 250px 70px,clip]{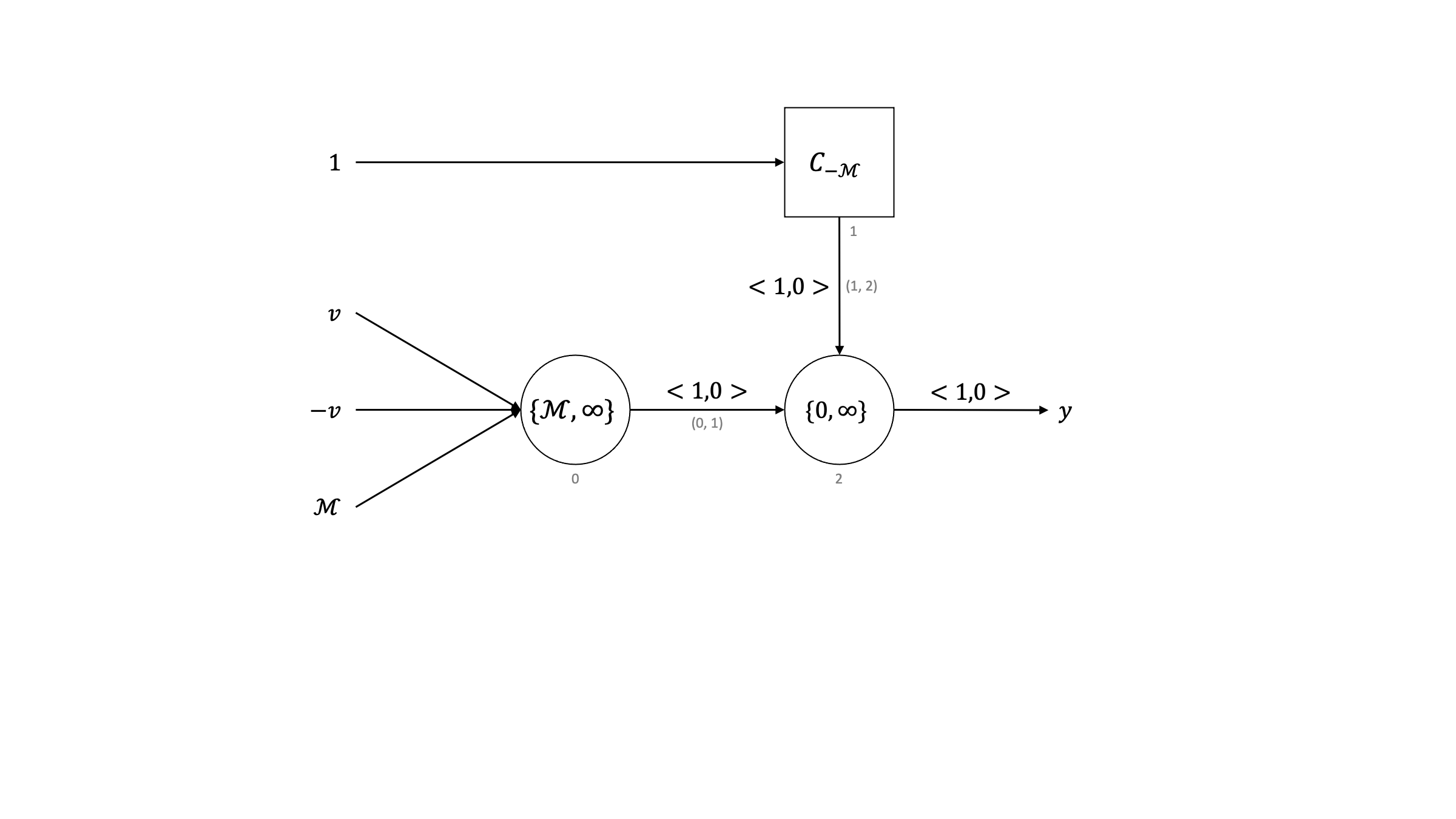}
        \caption{Trigger circuit.}
        \label{fig:trigger-mechanism-circuit}
    \end{subfigure}
    \begin{subfigure}[b]{0.4\textwidth}
        \centering
        \includegraphics[scale=0.4,trim=300px 200px 320px 220px,clip]{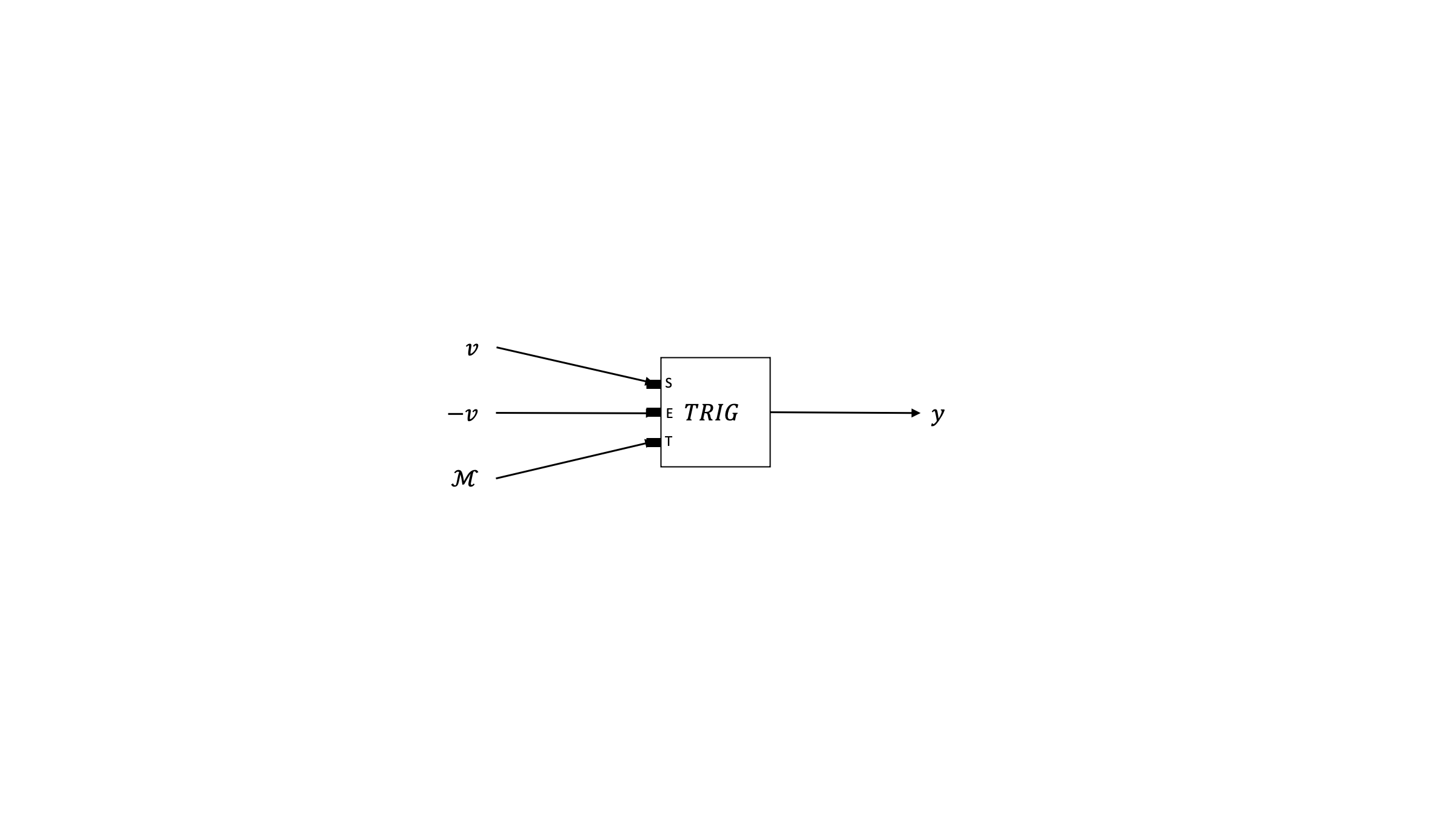}
        \caption{Trigger circuit abstraction.}
        \label{fig:trigger-mechanism-abstraction}
    \end{subfigure}
    \caption{Trigger circuit and its abstraction.}
    \label{fig:trigger-mechanism}
\end{figure*}



\begin{figure*}[h!]
    \centering
    \includegraphics[scale=0.57,trim=15px 0px 12px 2px,clip]{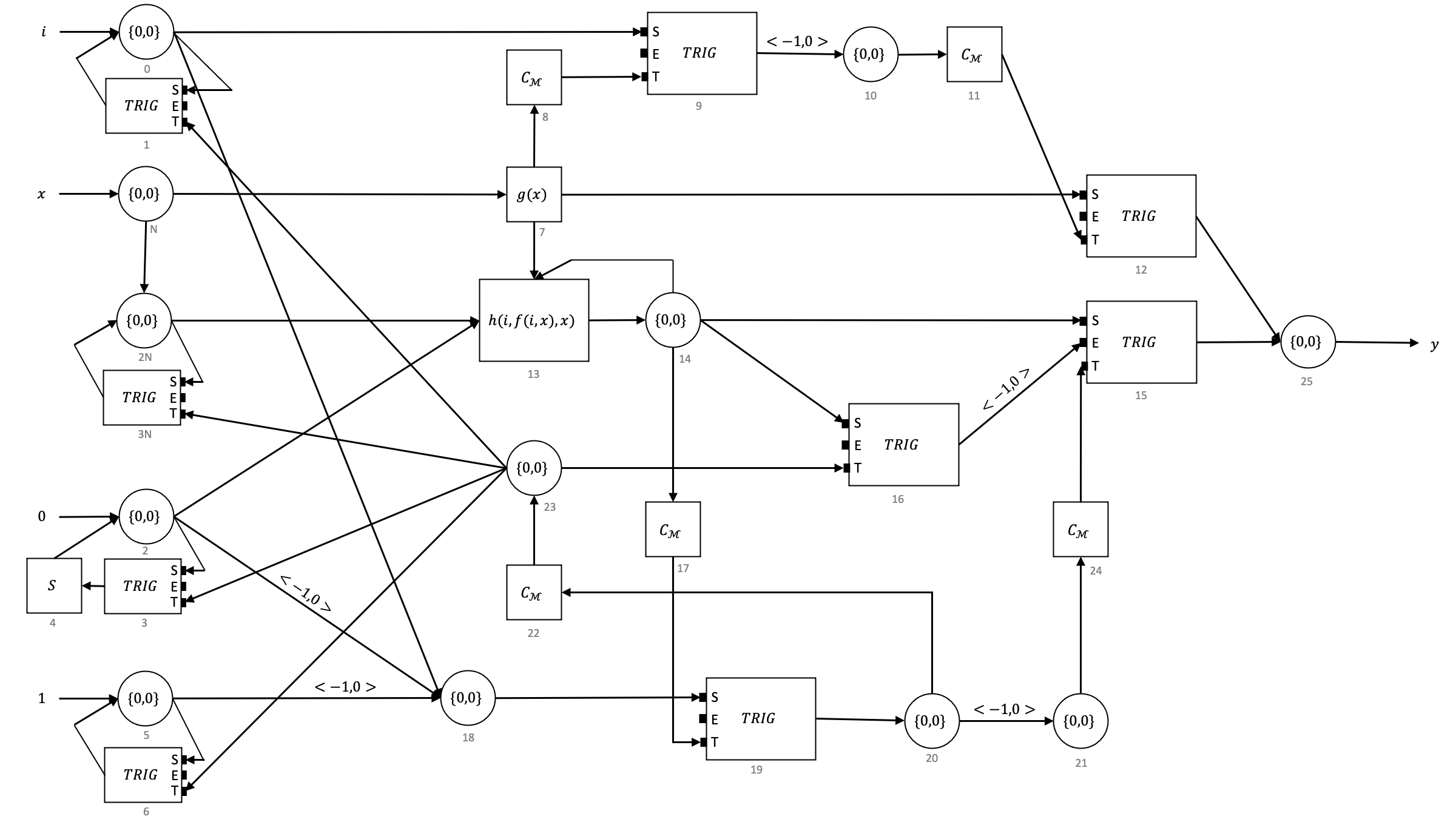}
    \caption{Neuromorphic circuit for the primitive recursion operator. All synapses are $\braket{1, 0}$ unless explicitly stated. References to all synapses can be inferred from the references of their pre-synaptic and post-synaptic circuit components. The input $x$ in neuron \texttt{N} actually represents all of $x_1, \ldots, x_N$ such as the one shown in Figure \ref{fig:minimization-operator-circuit}. We only show it as a single input for the sake of clarity in the figure.}
    \label{fig:primitive-recursion-operator-circuit}
\end{figure*}

\begin{figure}[t!]
    \centering
    \includegraphics[scale=0.4,trim=200px 190px 210px 190px,clip]{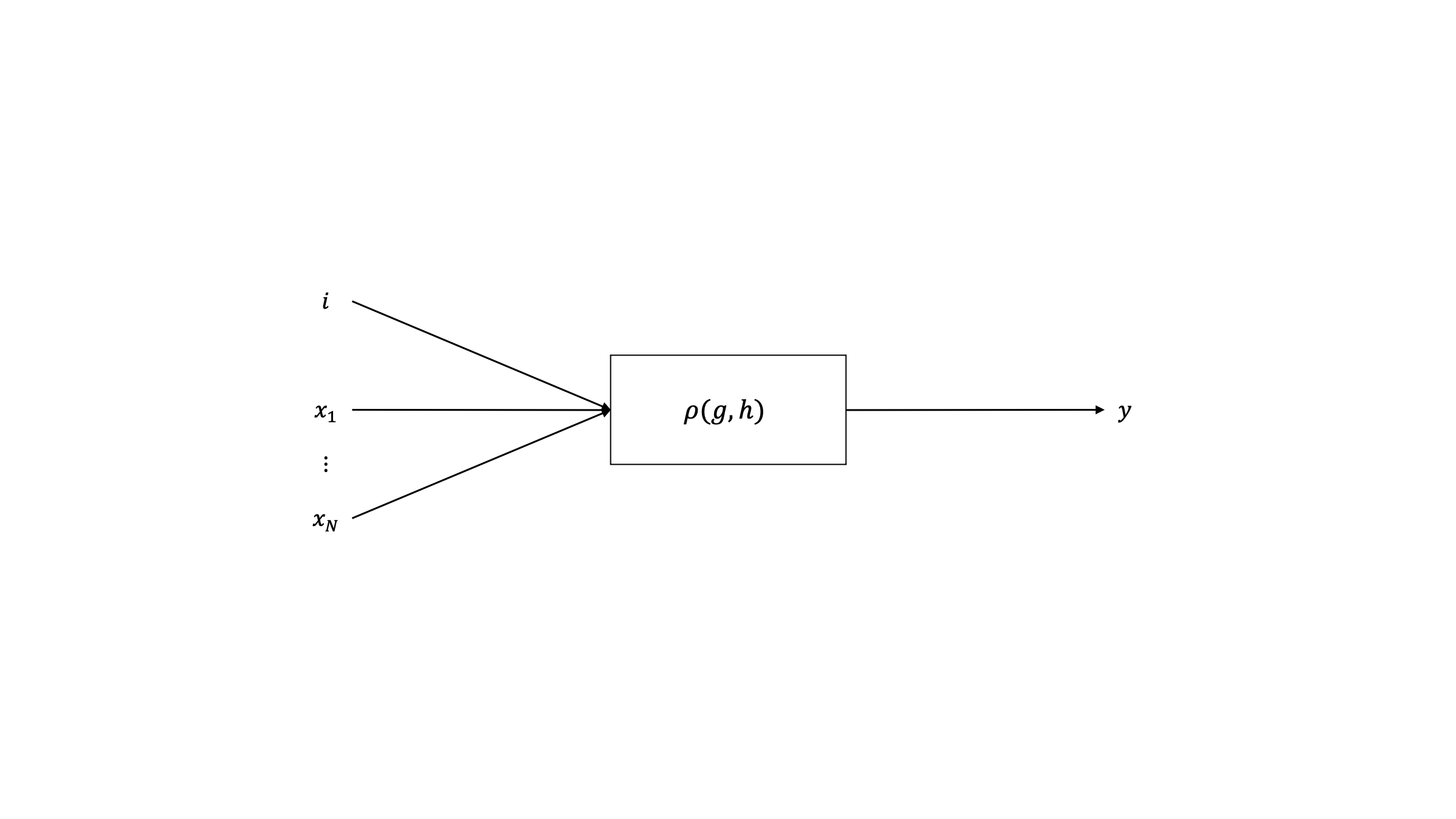}
    \caption{Abstraction for the primitive recursion operator.}
    \label{fig:primitive-recursion-operator-abstraction}
\end{figure}

Given the functions $g: \mathbb{N}^N \rightarrow \mathbb{N}$ and $h: \mathbb{N}^{N+2} \rightarrow \mathbb{N}$, the primitive recursion operator $\rho(g, h)$ equals the function $f$, defined as:
\begin{align}
    f(0, x_1, \ldots, x_N) &:= g(x_1, \ldots, x_N) \label{eq:recursion-base-case-definition} \\
    f(i+1, x_1, \ldots, x_N) &:= h(i, f(i, x_1, \ldots, x_N), x_1, \ldots x_N) \label{eq:recursion-general-case-definition}
\end{align}

Before describing the neuromorphic circuit for the primitive recursion operator, we describe the trigger circuit shown in Figure \ref{fig:trigger-mechanism-circuit}.
It stores a value $v$ in neuron \texttt{0}, which has a threshold of $\mathcal{M}$ and infinite leak.
The stored value can be erased by passing $-v$ to neuron \texttt{0}.
The trigger circuit returns $v$ stored in neuron \texttt{0} when triggered by an input of $\mathcal{M}$.
When this happens, neuron \texttt{0} spikes with a value of $\mathcal{M} + v$.
This value reaches neuron \texttt{2}, which has an internal state of $-\mathcal{M}$.
After receiving the signal from \texttt{(0, 1)}, its internal state becomes $v$ and it spikes.
The value $v$ is returned as the output of the trigger circuit.
It is important to note that the inputs $v$, $-v$ and $\mathcal{M}$ are not passed to the trigger circuit simultaneously.
The abstraction of the trigger circuit is shown in Figure \ref{fig:trigger-mechanism-abstraction}, which has three pins S, E and T, representing the store, erase and trigger functionalities.

Figures \ref{fig:primitive-recursion-operator-circuit} and \ref{fig:primitive-recursion-operator-abstraction} show the primitive recursion operator and its abstraction.
The inputs $i$ and $x$ are passed by the user.
Here, $x$ actually represents all of $x_1, \ldots, x_N$.
In an actual implementation, each of them would individually have the same connections as $x$, including the iterative mechanism shown by neurons \texttt{2N} and \texttt{3N}.
We omit showing $x_1, \dots, x_N$ for the sake of clarity of Figure \ref{fig:primitive-recursion-operator-circuit}.
In addition to $i$ and $x$, we also provide $0$ and $1$ to neurons \texttt{2} and \texttt{5}.
The $0$ sent to neuron \texttt{2} is referred to as $j$ and is sent as an input to the function $h$ shown in \texttt{13}. 
It acts as a proxy for $i$.
The trigger circuits \texttt{1} and \texttt{9}, and the neuron \texttt{18} receive $i$.
Trigger circuit \texttt{1} stores $i$ for the next iteration, and \texttt{9} stores $i$ in case we need to return $f(0, x_1, \ldots, x_N)$.
Neuron \texttt{18} is used to evaluate the the expression $i - j - 1$, which when equals zero, acts as the stopping criteria.
$x$ is sent to the function $g$ at \texttt{7}, and the neuron $2N$, from where, it is sent to the function $h$ at \texttt{13}.
$j$ is sent to trigger circuit \texttt{3} for the next iteration.
It is also sent to the function $h$ and neuron \texttt{18}, which evaluates the stopping criteria.
The input $1$ to neuron \texttt{5} is sent to the trigger circuit \texttt{6} for the next iteration.
It is also sent to neuron \texttt{18} for evaluating the stopping criteria.

In case $i$ equals $0$, we need to return $g(x)$ as the output of the entire circuit.
This is done in the branch \texttt{9}-\texttt{10}-\texttt{11}-\texttt{12}-\texttt{25}.
$i$ is stored in the trigger circuit \texttt{9}.
Parallely, $x$ is sent to $g$.
After $g(x)$ is evaluated, it is sent to \texttt{12}, where it waits to be returned.
It is also sent to $h$ and the constant function \texttt{8}, which in turn, triggers \texttt{9}.
Upon being triggered, \texttt{9} sends $i$ to neuron \texttt{10}, which checks if it is zero.
If $i$ equals zero, neuron \texttt{10} spikes.
Else, that branch of the circuit is not activated.
When neuron \texttt{10} spikes, \texttt{12} is triggered. 
It sends the value of $g(x)$ to neuron \texttt{25}, which is returned as desired.

If $i$ is greater than zero, we want to return the value of $h$ when $i-j-1$ equals zero.
$h$ receives $x$ from \texttt{2N}, $g(x)$ from \texttt{7} and $j$ from \texttt{2}.
Parallely, $i-j-1$ is evaluated in \texttt{18} and stored in \texttt{19}.
After $h$ is evaluated, it is sent to neuron \texttt{14}, which distributes it to the trigger circuits \texttt{15} and \texttt{16}, the constant function \texttt{17}, and $h$ itself in preparation for the next iteration.
The trigger circuit \texttt{19} is used to return the value of $h$ if necessary, while \texttt{16} is used to erase the value in \texttt{15}, if required.
The constant function \texttt{17} triggers \texttt{19}, which sends $i-j-1$ to \texttt{20}.
Neuron \texttt{21} checks if $i-j-1$ is zero.
Parallely, $i-j-1$ is also passed to \texttt{22}, which returns $\mathcal{M}$.
Neuron \texttt{23} distributes $\mathcal{M}$ to \texttt{1}, \texttt{3N}, \texttt{3}, \texttt{6} and \texttt{16}.
While the first four are used to initiate the next iteration, \texttt{16} is used to erase the value stored in \texttt{15}.
Note that the branch \texttt{23}-\texttt{3}-\texttt{4} increments $j$ for the next iteration.
In case $i-j-1$ equals zero, neuron \texttt{21} spikes.
Trigger circuit \texttt{15} is eventually triggered in this branch and the value of $h$ stored in it is returned as output of the entire circuit via neuron \texttt{25}.
In this case, although the branch \texttt{20}-\texttt{22}-\texttt{23}-\texttt{16} is active, it does not erase the value in \texttt{15} in time.
The signal in the branch \texttt{20}-\texttt{21}-\texttt{24} reaches \texttt{15} first and returns the output.
This is a consequence of our assumption that each synapse takes one unit of time to propagate a signal from its pre-synaptic to post-synaptic component.

Thus, when $i$ equals zero, the recursion operator returns $g(x)$ as required.
If $i$ is greater than zero, we keep incrementing $j$ and keep evaluating $h$ until $i-j-1$ equals zero.
At this point, the value of $h$ stored in \texttt{15} is returned as desired.

\subsection{Minimization Operator}
\label{sub:minimization-operator}
\begin{figure*}[t!]
    \centering
    \includegraphics[scale=0.6,trim=10px 75px 230px 25px,clip]{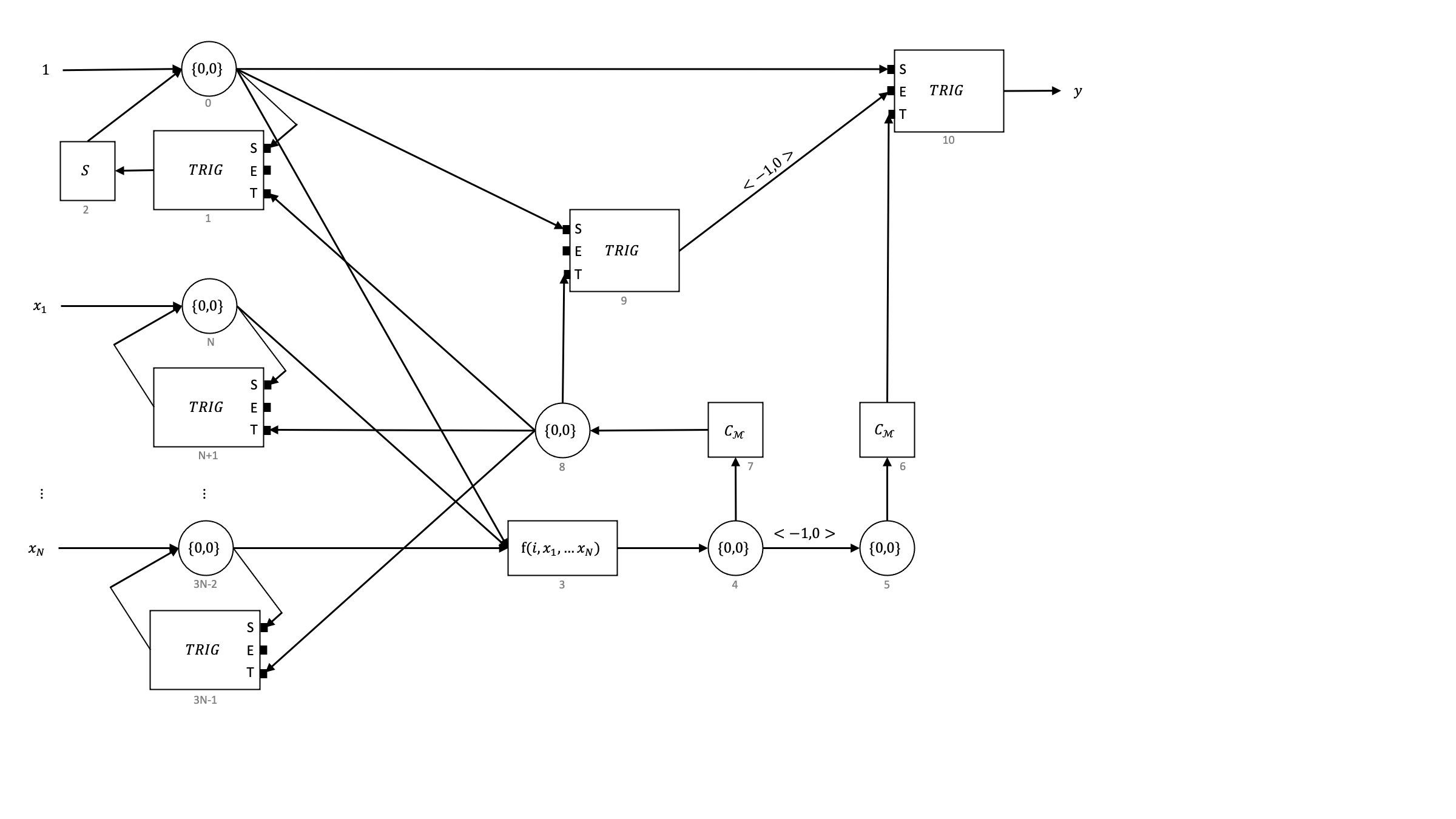}
    \caption{Neuromorphic circuit for the minimization operator. All synapses are $\braket{1, 0}$ unless explicitly stated. References to all synapses can be inferred from the references of their pre-synaptic and post-synaptic circuit components.}
    \label{fig:minimization-operator-circuit}
\end{figure*}

\begin{figure}[t!]
    \centering
    \includegraphics[scale=0.4,trim=200px 190px 210px 190px,clip]{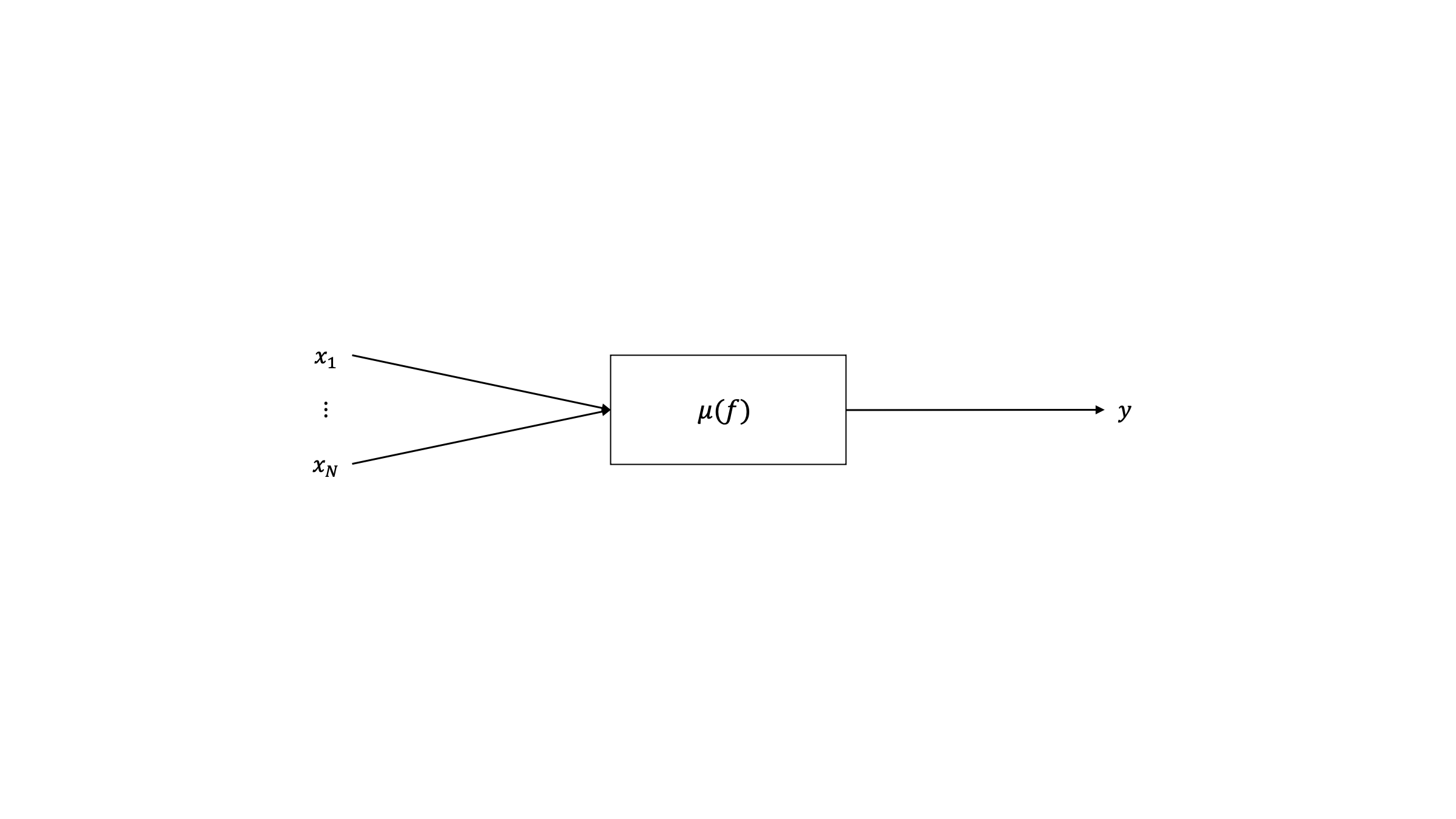}
    \caption{Abstraction for the minimization operator.}
    \label{fig:minimization-operator-abstraction}
\end{figure}

Given a function that only returns whole numbers for any natural number, we want to find the smallest natural number for which it returns zero.
Formally, given the function $f: \mathbb{N}^{N+1} \rightarrow \mathbb{N}$, the function $\mu(f)$ is defined as:
\begin{align}
    \mu(f)(x_1, \ldots, x_N) := z
    \label{eq:minimization-operator-definition}
\end{align}
such that:
\begin{align}
    f(i, x_1, \ldots, x_N) &> 0 \qquad \forall i = 1, 2, \ldots, z-1 \\
    f(z, x_1, \ldots, x_N) &= 0 
\end{align}

Figures \ref{fig:minimization-operator-circuit} and \ref{fig:minimization-operator-abstraction} show the minimization operator and its abstraction.
Neuron \texttt{0} receives $1$ and spikes.
Parallely, neurons \texttt{N}, \texttt{N + 2}, $\ldots$ \texttt{3N - 2} receive $x_1, x_2, \ldots, x_N$ and spike.
The value from neuron \texttt{0} is stored in the trigger circuits \texttt{1}, \texttt{9} and \texttt{10}.
\texttt{1} is used to increment the value of neuron \texttt{0} for the next iteration.
\texttt{10} stores and returns the value if and when needed.
\texttt{9} is used to erase the value stored \texttt{10} if required.
Values from neurons \texttt{N}, \texttt{N + 2}, $\ldots$ \texttt{3N - 2} are stored in the trigger circuits \texttt{N + 1}, \texttt{N + 3}, \ldots, \texttt{3N - 1} for the next iteration.
They are also sent to the function $f$ at \texttt{3}.

$f$ evaluates a value and sends it to neuron \texttt{4}, which spikes regardless of whether the value is zero or not.
If the value evaluated by $f$ is not zero, neuron \texttt{4} sends this value to the constant circuit \texttt{7}, which in turn, sends $\mathcal{M}$ to neuron \texttt{8}.
Neuron \texttt{8} distributes $\mathcal{M}$ to the trigger circuits \texttt{9}, \texttt{1}, and \texttt{N + 1} through \texttt{3N - 1}.
Value stored in \texttt{9} is used to erase the value stored in \texttt{10}.
The trigger circuits \texttt{1} and \texttt{N + 1} through \texttt{3N - 1} initiate the next iteration.
The outgoing value from \texttt{1} is incremented by the successor function \texttt{2} for the next iteration.
The other branch coming out of neuron \texttt{4} is not activated as neuron \texttt{5} does not spike.

If $f$ returns zero, then neuron \texttt{5} spikes and sends this value to the constant function \texttt{6}.
It sends $\mathcal{M}$ and triggers \texttt{10}, which returns the  value stored in it.
The value returned by \texttt{10} is also the output $y$ of the entire circuit, and equals the smallest natural number for which $f$ returns $0$.
In this case, the branch \texttt{4}-\texttt{7}-\texttt{8}-\texttt{9}-\texttt{10} is also active.
However, \texttt{10}  still returns the correct output before its stored value is erased because the signal coming from the branch \texttt{4}-\texttt{5}-\texttt{6} reaches \texttt{10} first in initiates return of its stored value.

\subsection{Discussion}
\label{sub:discussion}


We show in this paper that the model of neuromorphic computing defined in Section \ref{sec:model} is at least as strong as the Turing machine, therefore Turing-complete.
Furthermore, since our model is simple, any neuromorphic implementation that can at least realize this model would also be Turing-complete.
The neuromorphic functionality provided by our model can potentially be leveraged to prove the Turing-completeness of other models of computation.
We keep synaptic delay as a parameter in our model of computing, but never use it in the proof in Section \ref{sec:proof}.
So, neuromorphic computing is Turing-complete even without the synaptic delay.
However, we keep it in our model for the sake of convenience as it is used in many spatio-temporal encoding schemes as well as several neuromorphic algorithms.
An implicit assumption that we make about all boxed functions in all our circuit diagrams is that they can be computed using our model of neuromorphic computing.
Finally, through this paper, we wish to confirm the Turing-completeness of our model of neuromorphic computing.
The objective of this paper is not to indicate any broader implications for neuromorphic computing or even computing in general.


\section{Practical Considerations}
\label{sec:practical-considerations}

In defining our model of computation in Section \ref{sec:model}, we assumed the existence of infinitely many neurons and synapses enabling all-to-all-connectivity. 
In a physical realization, we expect to have a finite number of neurons, each having a finite number of incoming and outgoing synapses.
We observe a similar relation between the Turing machine and the von Neumann architecture.
While the Turing machine assumes the existence of an infinite tape, the von Neumann architecture has finite memory in practice.
We also assumed that a neuron could be connected to itself. 
If this functionality is not offered on hardware, it can be realized by using a proxy neuron that sends the signal back to the original neuron. 

We have established that neuromorphic computing is capable of general-purpose computing by showing it can compute $\mu$-recursive functions and operators.  
Systems of spiking neurons and synapses can also be used to implement Boolean logic functions such as AND, OR, and NOT.  
Though it is possible to implement computation in these ways on a  neuromorphic system, it may not be the best way.  
There may be better ways to realize complex functionality with neurons and synapses than defining computation in the same way as it is defined for traditional computers. 
As we are understand how to leverage the computational capabilities of neuromorphic computers to the fullest, we must also learn how to program them effectively and practically. 
As the notion of what computers look like continues to evolve with the end of Moore's law, many more ways of defining how computation is performed will emerge.

\section{Conclusion}
\label{sec:conclusion}
We propose a simple model of neuromorphic computing and prove that it is Turing-complete.
With this proof, we establish that neuromorphic systems are capable of general-purpose computing and can operate as independent processors, not just as a co-processor in a larger system that is driven by a traditional CPU.  
Compounding the general-purpose computability of neuromorphic computers with their extremely low power nature, there may be an opportunity to implement increasingly large amounts of computation on a neuromorphic computer---this will result in significant energy savings.  
With the explosion of computing devices from the cloud and high performance computing, all the way to smart phones and wearable technologies, even small power savings that can be obtained through implementation of common algorithms on more efficient hardware can have a significant impact on the carbon footprint of computing.


\section*{Acknowledgment}

This manuscript has been authored in part by UT-Battelle, LLC under Contract No. DE-AC05-00OR22725 with the U.S. Department of Energy. The United States Government retains and the publisher, by accepting the article for publication, acknowledges that the United States Government retains a non-exclusive, paid-up, irrevocable, world-wide license to publish or reproduce the published form of this manuscript, or allow others to do so, for United States Government purposes. The Department of Energy will provide public access to these results of federally sponsored research in accordance with the DOE Public Access Plan (http://energy.gov/downloads/doe-public-access-plan).
This work was funded in part by the DOE Office of Science, Advanced Scientific Computing Research (ASCR) program.
This material is based upon work supported by the U.S. Department of Energy, Office of Science, Office of Advanced Scientific Computing Research, Robinson Pino, program manager, under contract number DE-AC05-00OR22725.

\section{Author Contributions}
P.D. developed the model of neuromorphic computing, devised neuromorphic circuits to compute all $\mu$-recursive functions and operators, and wrote majority of the manuscript. 
C.S. and B.K. helped in developing the model of neuromorphic computing, validated the proof of Turing-completeness, and helped in writing the paper.
T.P. was the principal investigator on the U.S. Department of Energy project that funded this work.

\section{Competing Interests}
The authors declare no competing interests.

\bibliographystyle{unsrt}  


\bibliography{references}






\end{document}